%% file: main_arxiv.tex
\documentclass[10pt,twocolumn,letterpaper]{article}

\usepackage[pagenumbers]{cvpr}


\definecolor{cvprblue}{rgb}{0.21,0.49,0.74}
\usepackage[pagebackref,breaklinks,colorlinks,allcolors=cvprblue]{hyperref}
\usepackage{xcolor}
\usepackage{xspace}
\usepackage[linesnumbered,ruled,vlined]{algorithm2e}
\usepackage{url}
\usepackage{xspace}
\usepackage[hypcap=false]{caption}
\usepackage{graphicx}
\usepackage{booktabs}
\usepackage{arydshln}
\usepackage{multirow}
\usepackage{etoolbox,lipsum}
\usepackage{array}
\usepackage{supertabular}
\usepackage{makecell}
\usepackage{tabularx}
\usepackage{xcolor,colortbl}
\usepackage[most]{tcolorbox}
\usepackage{fancyhdr}

\newcommand{\modelname}{Latent-Reframe\xspace}

\title{Latent-Reframe: Enabling Camera Control for \\Video Diffusion Model without Training}

\author{Zhenghong Zhou\textsuperscript{*}, ~~Jie An\textsuperscript{*}, ~~ Jiebo Luo\\
University of Rochester\\
{\tt\small zzhou72@ur.rochester.edu, \{jan6,jluo\}@cs.rochester.edu}
}

\begin{document}
\twocolumn[{
\maketitle
\renewcommand\twocolumn[1][]{#1}
    \begin{center}
        \centering
        \includegraphics[width=1.0\linewidth]{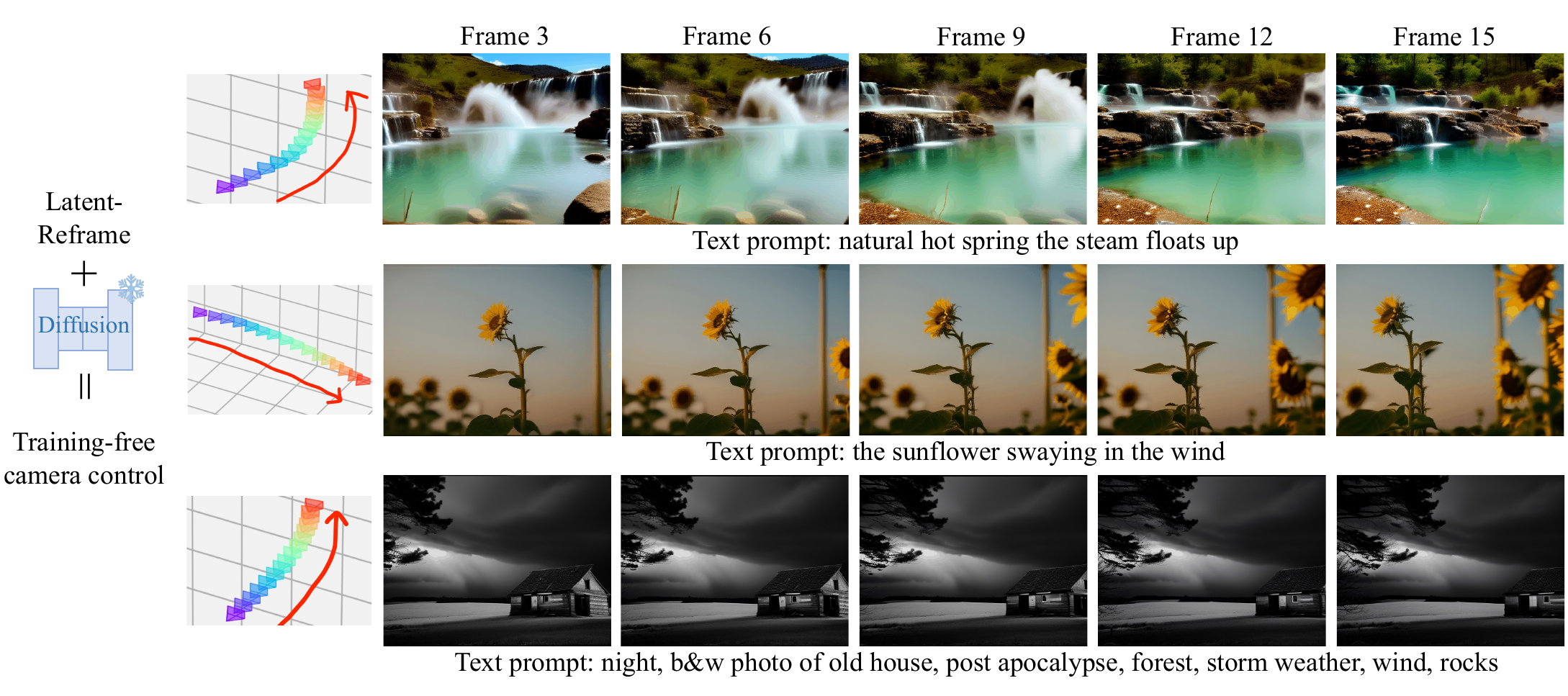}
        \vspace{-7mm}
        \captionof{figure}{The proposed \modelname enables a text-to-video diffusion model to generate high-quality videos that accurately follow both the text prompt and a specified camera trajectory, all without requiring additional training.}
        \label{fig:teaser}
    \end{center}
}]

\renewcommand{\thefootnote}{\fnsymbol{footnote}}
\footnotetext[1]{Equal contribution}
\input{sec/0_abstract}    
\input{sec/1_intro}
\input{sec/2_relatedwork}
\input{sec/3_method}
\input{sec/4_experiment}
\input{sec/5_conclusion}
\clearpage
{
    \small
    \bibliographystyle{ieeenat_fullname}
    \bibliography{main}
}

\input{sec/6_supp}

\end{document}

%% file: sec/0_abstract.tex
\begin{abstract}
Precise camera pose control is crucial for video generation with diffusion models. Existing methods require fine-tuning with additional datasets containing paired videos and camera pose annotations, which are both data-intensive and computationally costly, and can disrupt the pre-trained model’s distribution.
We introduce \modelname, which enables camera control in a pre-trained video diffusion model without fine-tuning. Unlike existing methods, \modelname operates during the sampling stage, maintaining efficiency while preserving the original model distribution.
Our approach reframes the latent code of video frames to align with the input camera trajectory through time-aware point clouds. Latent code inpainting and harmonization then refine the model’s latent space, ensuring high-quality video generation.
Experimental results demonstrate that \modelname achieves comparable or superior camera control precision and video quality to training-based methods, without the need for fine-tuning on additional datasets.
\end{abstract}

%% file: sec/1_intro.tex
\section{Introduction}
\label{sec:intro}

Generating a video that follows a user-defined camera trajectory is a key task for controllable video generation.
Existing methods, such as CameraCtrl~\citep{cameractrl} and MotionCtrl~\citep{motionctrl}, inject camera control into a pre-trained video diffusion model by fine-tuning it with an additional dataset containing paired videos and camera trajectories.

While these methods demonstrate good performance, they face two key challenges: 1) Fine-tuning requires an extra dataset annotated with camera trajectories, which is labor-intensive to collect and computationally expensive in training, even with techniques like Low-Rank Adaptation (LoRA)~\citep{lora}. 2) Fine-tuning can disrupt the learned distribution of the pre-trained video diffusion model, potentially degrading the quality of the generated videos if the fine-tuning dataset is of lower quality.
This raises an important question: Can camera control be injected into a pre-trained video diffusion model without encountering these issues?

Building on insights from image editing research~\citep{shi2024dragdiffusion,zhang2024gooddrag} showing that pre-trained diffusion models can accommodate latent code manipulation, we introduce \modelname, a sampling-stage method that enables camera control in video diffusion models without fine-tuning. Midway through the denoising process, our method performs two key steps: latent code reframing to align the video latent with the target camera pose and latent space rehabilitation to inpaint and harmonize the latent code.

Given a pre-trained video diffusion model, such as AnimateDiff~\citep{animatediff}, we use MonSt3R~\citep{monst3r} to extract per-frame 3D point clouds and corresponding camera poses from partially denoised latent codes. We then adjust each frame’s camera pose to match the user-defined target positions and reproject the time-aware 3D point clouds to video frames based on these adjusted poses, creating frames that follow the desired camera trajectory. These reframed frames are then converted to latent codes via the VAE encoder, ready for the remaining denoising steps.

The latent code reframing process can create blank regions in the video’s latent code due to occlusion. To address this, we propose a latent space rehabilitation process to inpaint these regions and harmonize the latent code. Specifically, we add noise to both the warped latent codes and blank regions, then denoise to fill in the missing content. Inspired by FIFO-Diffusion~\citep{fifo}, we apply lighter noise to the warped latent codes, enabling the blank regions to incorporate content from the reframed area, which improves video content preservation. After rehabilitation, we complete the remaining denoising steps, producing a video aligned with the desired camera trajectory.

We compare \modelname with state-of-the-art, training-based baselines, including CameraCtrl~\citep{cameractrl} and MotionCtrl~\citep{motionctrl}, using $10$ prompts and $80$ camera trajectories. Experimental results show that our method, without fine-tuning, enables a pre-trained video diffusion model to achieve comparable or even superior camera control precision. Additionally, videos generated by our method demonstrate higher quality, as measured by FID~\citep{FID} and FVD~\citep{fvd} scores, indicating a stronger capacity to preserve the video generation quality of the pre-trained model.

This work contributes to the field by introducing \modelname, which leverages latent reframing through 3D point cloud remapping, enabling camera control in a pre-trained video diffusion model without fine-tuning. Results show that \modelname produces videos with precise camera pose trajectories and superior quality compared to training-based methods, highlighting the potential of 3D information to enhance 2D video generation.

%% file: sec/2_relatedwork.tex
\section{Related work}
\label{sec:relatedwork}

\paragraph{Camera Control for Video Diffusion Model.}
Recent advancements in diffusion models~\citep{sohl2015deep,song2020score,ho2020denoising,kadkhodaie2020solving,nichol2021improved,an2024bring,an2024inductive} have enabled the generation of high-quality videos. To make video diffusion models~\citep{ho2022vdm,ho2022imagen,voleti2022mcvd,singer2022make,villegas2022phenaki,2023videocomposer,yin2023dragnuwa,guo2023animatediff,chen2024videocrafter2,an2023latent,blattmann2023videoldm,2023pyoco,2023tune,chen2023videocrafter1,chen2024videocrafter2} follow a camera pose trajectory, two types of approaches are commonly used.

The first approach is implicit camera pose control, where poses are embedded and injected into the UNet denoising network. Methods like MotionCtrl~\cite{motionctrl}, CameraCtrl~\cite{cameractrl}, CamCo~\cite{camco}, and Direct-a-Video~\cite{direct} fall under this category. However, these methods require extensive training on large, high-quality datasets, demanding significant computational and data resources.

The second approach uses explicit pose control, where a point cloud is extracted, and the pose is used to render frames before denoising. \modelname aligns with this category, similar to CamTrol~\cite{CamTrol} and ReCapture~\cite{ReCapture}. Unlike these methods, which rely on static point cloud rendering and are limited to image-to-video generation, \modelname employs time-aware point clouds, enabling both text-to-video and image-to-video generation. Empirically, time-aware point clouds outperform static ones in capturing complex video dynamics, making \modelname more versatile and effective for diverse tasks.

\paragraph{3D Reconstruction and Novel View Synthesis.} 
3D reconstruction and novel view synthesis aim at learning scene details from sparse input views~\cite{nerf, 3Dgs}. DUSt3R~\cite{dust3r} advances 3D reconstruction from sparse inputs by enabling scene point cloud reconstruction and camera pose estimation from just two images, while MonST3R~\cite{monst3r} further improves results in dynamic reconstruction. Other methods, such as InstanSplat~\cite{instantsplat}, NoPoSplat~\cite{noposplat}, and Splatt3R~\cite{splatt3r}, support 3D Gaussian Splatting reconstruction from extremely sparse views. Integrated with video diffusion models, tools like ViewCrafter~\cite{viewcrafter} and ReconX~\cite{reconx} have achieved enhanced 3D and even 4D reconstruction.
These advances demonstrate that 3D reconstruction models can infer dense scene information from sparse data, providing strong potential for generative tasks. However, current methods typically require extensive training, especially with video diffusion models, which are computationally intensive. This paper proposes a method to enhance generative model quality and control using 3D reconstruction without additional training.

%% file: sec/3_method.tex
\begin{figure*}[t]
    \centering
    \includegraphics[width=0.95\linewidth]{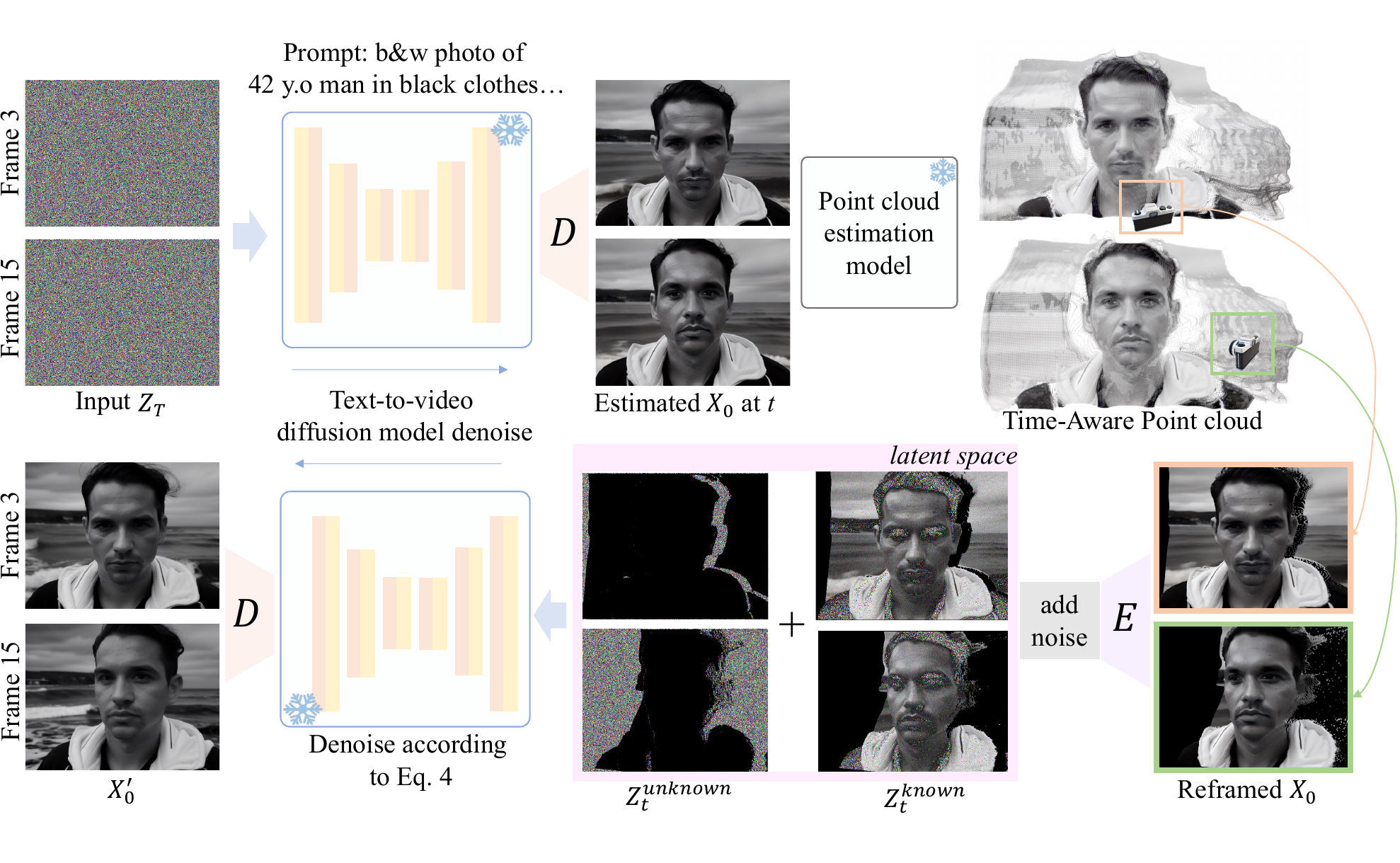}
    \vspace{-5mm}
    \caption{\textbf{Overview of the proposed \modelname.} In the middle of the denoising process of a pre-trained video diffusion model, we first extract time-aware 3D point cloud via a point cloud estimation model, which takes $x_0$ estimated by the halfway denoised latent code as the input. Next we reframe $x_0$ according to the target camera pose and the per-frame point cloud. Then we use the proposed latent space rehabilitation approach to inpaint the blank region due to occlusion and harmonize the latent code. $x_0^\prime$ is the reframed video after finishing the remaining denoising steps, which follows the target camera pose trajectory.}
    \label{fig:method}
\end{figure*}

\section{Method}

To enable camera control in a video diffusion model through latent code manipulation without training, we face two key challenges: 1) Extracting precise camera poses from partially denoised latent codes and reframing the latent codes to align with the target camera pose trajectory. 2) Inpainting and harmonizing the reframed video latents to preserve the quality of the generated videos. 

This section details our solutions to these challenges. In Sec.\ref{method:preli}, we outline the preliminaries of video diffusion models. Sec.\ref{method:render} explains our approach to manipulating latent codes during denoising to align videos with the desired camera trajectory. Sec.~\ref{method:inpaint} describes how we rehabilitate the manipulated latent codes to maintain video quality. Fig.~\ref{fig:method} and Alg.~\ref{alg:algorithm} provide an overview and a pseudo-code of the proposed method, respectively.

\subsection{Preliminary of Video Diffusion Model}
\label{method:preli}

A video diffusion model maps the training video distribution to a Gaussian noise distribution through a forward and backward process. The forward process adds noise to the video, while the backward process iteratively recovers a clean video from random Gaussian noise. For latent video diffusion models, the training data consists of latent codes $z_0=\{ z_0^0, z_0^1,...,z_0^F\}$ where $z_0^n$ represents the latent code of the $n^{\rm th}$ video frame, and $F$ is the number of frames in a video. During the forward process, noise is added to $z_0$ as follows:
\begin{equation}
    z_t = \sqrt{\bar{\alpha}_t}z_0 + \sqrt{1 - \bar{\alpha}_t}\epsilon,\ \ \epsilon\sim\mathcal{N}\left(\mathbf{0}, \mathbf{I}\right)
    \label{eq:2}
\end{equation}
where $z_t$ is the noisy latent code at diffusion step $t$. $\alpha_t$ and $\bar{\alpha}_t$ are predefined constants by diffusion model.
Eq.~\ref{eq:2} allows recovery of the clean latent code $z_0$ from partially denoised $z_t$, enabling the extraction of 3D point clouds and camera poses from halfway-denoised video latents (see Sec.~\ref{method:render} for details). 
The backward process takes a noisy latent $z_t$ and produces the one-step denoised latent $z_{t-1}$. In this work, we use DDIM~\cite{ddim} sampler for the backward process.

The training objective of a video diffusion model is
\begin{equation}
    \mathcal{L} = \mathbb{E}_{z_0,\epsilon,t}\left[\left\|\epsilon-\epsilon_\theta(z_t,t,c)\right\|_2^2\right],
    \label{eq:3}
\end{equation}
where $c$ is the optional conditions such as text embedding for text-to-video generation. $\theta$ denotes the network architecture (\eg,  UNet~\citep{unet} and DiT~\citep{dit}), which captures spatial and temporal information to predict the added noise $\epsilon$.

\subsection{Latent Reframing} 

\label{method:render}

We manipulate the latent code midway through the denoising process to ensure the generated video follows the specified camera pose trajectory. Starting from random noise at diffusion step $T$, when the diffusion model reaches a chosen denoising step $t$, we first derive an approximate $z_0$ from $z_t$ using Eq.~\ref{eq:2} and then reconstruct the pixel-space video $x_0$ based on this estimate of $z_0$. Next, we use MonST3R~\cite{monst3r} to transform the 2D video into a sequence of time-aware 3D point clouds, where each point cloud corresponds to one video frame.

Notably, MonST3R requires image pairs as input, while the video contains $16$ frames. Inferring each pair sequentially would lead to scale misalignment in point maps, necessitating a global alignment.

Following DUSt3R~\cite{dust3r}, we construct a connectivity graph $G(V, \mathcal{E})$, where the $16$ images form the vertices $V$, and each edge $e = (n, m) \in \mathcal{E}$ represents an image pair $(I_n, I_m)$ for model processing. To optimize computation, we build this graph using a sliding window approach, pairing images within each window. For example, with a window size of $3$, we generate $6$ image pairs, sliding the window until all frames are processed.

Our objective is to obtain globally aligned point maps for each frame in a common coordinate system, denoted as ${P_n \in \mathbb{R}^{W \times H \times 3}}$ for all images $n = 1, \ldots, 16$. Each image pair in the connectivity graph, when processed by MonST3R, produces a pair of point clouds. These point clouds can be mapped to the coordinate system of the globally aligned point maps by applying a scaling factor $s$ and a pose transformation $\tau \in \mathbb{R}^{3 \times 4}$. We aim for these transformed point clouds to align with the globally aligned points.
To achieve this, the globally aligned point maps $P$, scale $s$, and pose $\tau$ are defined as learnable parameters, framing the task as an optimization problem for backpropagation. The objective function is:
\begin{equation}
P^*=\underset{P, \tau, s}{\arg \min } \sum_{e \in \mathcal{E}} \sum_{v \in e} \sum_{i=1}^{H W} C_i^{v, e}\left\|P_i^v-s_e \tau_e Q_i^{v, e}\right\| ,
\end{equation}
where $C_i^{v, e}$ and $Q_i^{v, e}$ represent the confidence score and point map, respectively, for the $i$-th pixel in image $v$ on edge $e$. Here, $H$ and $W$ denote the height and width of frames.

Upon completion of the optimization, we obtain the point maps and corresponding poses for the $16$ video frames in a unified coordinate system. Subsequently, we perform camera pose shifting to determine the target new poses for the image sequence. A series of poses from RealEstate10K~\cite{realestate} is utilized to compute the target poses. Initially, we process these poses by calculating the relative pose, which involves determining the changes in pose relative to the first frame. After this, we apply normalization and scale adjustment. Finally, these processed poses are multiplied by the original poses of 16 frames, resulting in the target new poses for video frames.

Besides lifting each video frame into a time-aware 3D point cloud, one can alternatively reconstruct a global point cloud with all video frames~\citep{CamTrol}, we empirically find that our time-aware point cloud results in fewer artifacts and better video quality when performing latent reframing. Please refer to Sec.\ref{exp:ablation} for more details.

\subsection{Latent Rehabilitation}
\label{method:inpaint}

Due to the presence of occlusions, the reframed video frames $x_t^\prime$ and their latent codes $z_t^\prime$ subjected to camera pose shifting often exhibit blank regions, resulting in reduced image quality. To address this, we propose a simple yet effective method to rehabilitate these occluded regions without using external models.

\paragraph{Latent Code Inpainting.} Inspired by RePaint~\cite{repaint}, we use the video diffusion model itself to inpaint blank regions in the video latent code. After obtaining the reframed video latent code $z_t^\prime$ at the $t$-th denoising step, rather than proceeding with further denoising, we reintroduce noise to $z_t^\prime$.

Specifically, we create a mask $m$ to identify occluded regions lacking valid pixel values by detecting points unrendered by the time-aware point cloud. Random Gaussian noise is added to the masked regions, reverting them to the initial state of the denoising process at time step $T$. Noise is also added to the unmasked regions; however, during the follow-up denoising step, only the masked regions are updated, while the unmasked regions are directly computed with Eq.~\ref{eq:2} using $z_0^\prime$. This iterative process continues until the $t$-th denoising step, by which point the noise levels in both occluded and non-occluded regions become aligned.
\begin{equation}
\begin{aligned}
& z_{t-1}^{\text {known }} \sim \sqrt{\bar{\alpha}_{t-1}}z_0^\prime + \sqrt{1 - \bar{\alpha}_{t-1}}\epsilon,\ \ \epsilon\sim\mathcal{N}\left(\mathbf{0}, \mathbf{I}\right) \\
&z_{t-1}^{\text {unknown }}  \sim \text{DDIM}\left(z_t^\prime, t\right), \\
&z_{t-1}^\prime =m \odot z_{t-1}^{\text {known }} +(1-m) \odot z_{t-1}^{\text {unknown }}
\end{aligned}
\label{eq:sample}
\end{equation}
We then finish the remaining denoising steps with entire latent codes uniformly.

\paragraph{Latent Space Harmonization.} Directly applying the inpainting method to blank, unknown regions can lead to disharmonious video content. This is likely because, at a denoising step $t$, the content in the unknown regions doesn’t fully incorporate the video content in the known regions, creating inconsistencies.

Inspired by FIFO-diffusion~\cite{fifo}, our approach leverages content from known regions with reduced noise levels to guide the generation of unknown regions with higher noise levels, ensuring consistency. Specifically, at the $t$-th denoising step, the known regions are assigned noise at the $(t-3)$ level, while the unknown regions retain noise at the $t$ level. Ablation studies support the effectiveness of this approach; for further details, see Sec.~\ref{exp:ablation}.

\begin{algorithm}[t]
\caption{The denoising process of a video diffusion model using \modelname.}
\label{alg:algorithm}

\KwIn{Initial random noise for $16$ frames: $\{z_T^i\}_{i=1}^{16} \sim \mathcal{N}(0, \mathbf{I})$, Constant parameters of noise scheduler: $\{\alpha_t\}_{t=1}^T$ and $\{\bar{\alpha}_t\}_{t=1}^T$, Pre-trained video diffusion backbone: $\epsilon_\theta$,}
\KwIn{diffusion step to perform \modelname $t_w$, point cloud estimation model: $\mathcal{R}$}
\KwOut{Latent code the produced video: $\{z_0^i\}_{i=1}^{16}$. }

\For{$t = T$ \KwTo $1$}{
    $\epsilon \gets \epsilon_\theta(z_t, t)$ 
    
    \eIf{$t == t_w$}{
        Estimated $z_0 \gets z_{t}$ using Eq.~\ref{eq:2}
        
        Estimated $x_0 \gets z_0$ using VAE decoder
        
        Time-aware point cloud $P_{i=1}^{16}\gets \mathcal{R}(x_0)$
        
        Reframed $x_0^\prime \gets (x_0, P_{i=1}^{16})$
        
        Reframed $z_0^\prime\gets x_0^\prime$ using VAE encoder

        Reframed $z_T^\prime \gets z_0^\prime$ with reduced noise level for the known region
        
         \For{$t^\prime = T$ \KwTo $t_w$}{
            $z_{t-1}^\prime \gets z_t^\prime$ using Eq.~\ref{eq:sample}
         }
        
    }{
        $z_{t-1} \gets z_t$ using DDIM~\citep{ddim} sampler 
    }
}

\Return $z_0$
\end{algorithm}

%% file: sec/4_experiment.tex
\section{Experiment}

In this section, we evaluate our method through comparative and ablation experiments. Section~\ref{exp:setting} details the implementation specifics. Section~\ref{exp:mainresult} compares our approach with other training-based method enabling camera control of video generation methods, providing both quantitative and visual analyses. Section~\ref{exp:ablation} examines the impact of individual design components of our method.

\subsection{Experiment Settings}
\label{exp:setting}

\textbf{Implementation Details.}
We choose AnimateDiff~\cite{animatediff} as our base model, the generated videos contain $16$ frames with spatial resolution $512\times384$. We employed $25$ DDIM denoising steps. For a fair comparison, results from MotionCtrl and CameraCtrl using AnimateDiff are also presented.  Point maps were generated using MonST3R~\cite{monst3r}, and rendering was performed with PyTorch3D~\cite{pytorch3d}. Inference was conducted on a single NVIDIA RTX A5000 GPU, with each sample requiring approximately $3$ minutes. Our method does not require training and can be readily adapted to the inference steps of any other diffusion other models.

\begin{figure*}[t]
    \centering
    \includegraphics[width=0.95\linewidth]{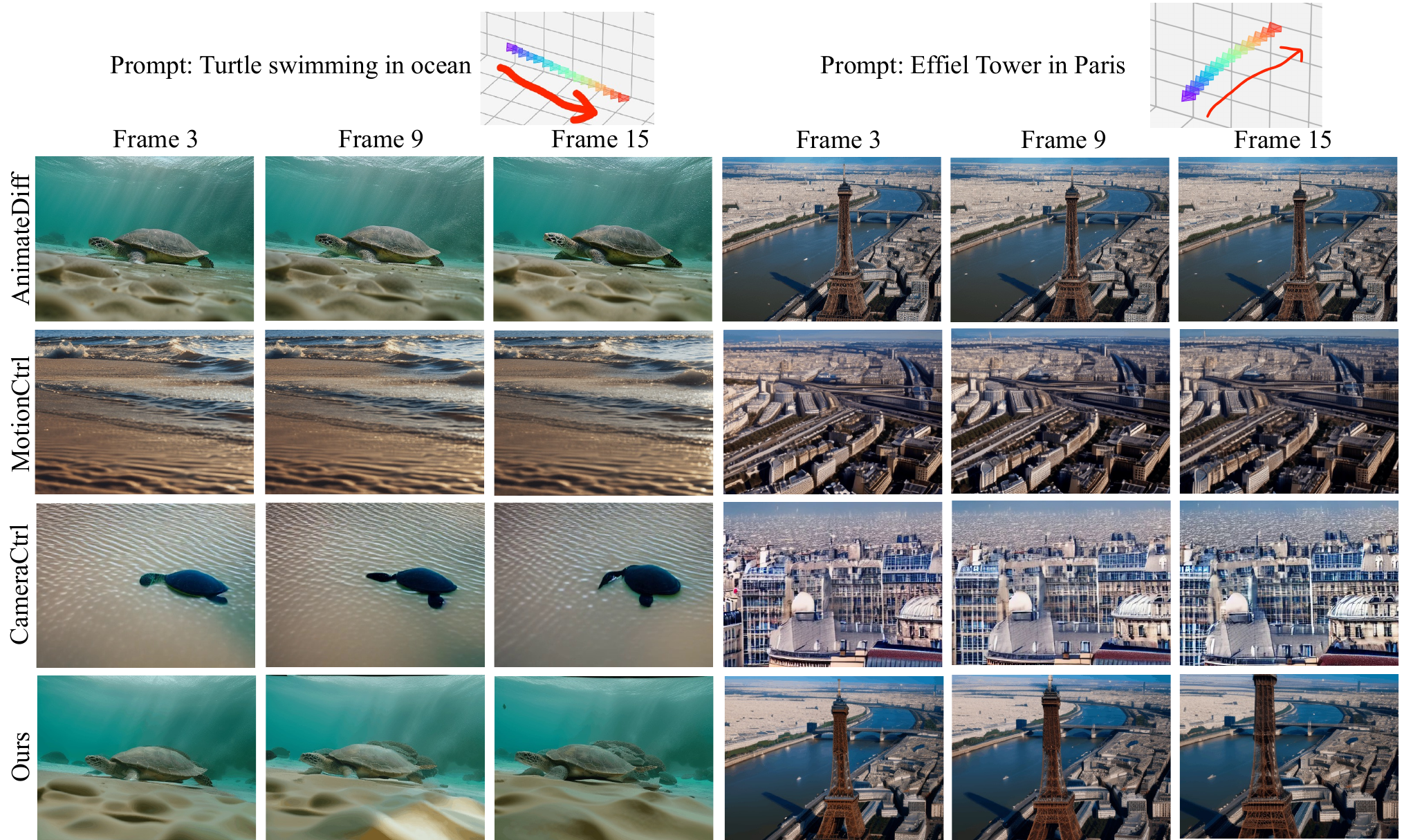}
    \caption{\textbf{Visual comparison with state-of-the-art methods.} The proposed \modelname can generate videos following the given camera trajectory without training. The video quality and the camera pose accuracy are comparable with the compared training-based methods. AnimateDiff is the pre-trained text-to-video diffusion model used by all the compared method. Only \modelname can keep the learned video distribution of AnimateDiff.}
    \label{fig:comparison}
\end{figure*}

\noindent \textbf{Evaluation Details.}
Regarding the evaluation dataset, we utilize $10$ prompts mentioned in MotionCtrl and CameraCtrl, combined with $80$ poses from the RealEstate10K dataset~\cite{realestate}, enabling the generation of $800$ videos. Given that both MotionCtrl and CameraCtrl employ implicit poses for embedding, while our method uses explicit poses for rendering, aligning the pose scales, such as translation scale, of videos produced by different methods poses a significant challenge, due to the uncertain scale of the generated scenes. 
We adjust the camera motion scale for each tested pose manually to ensure that the movements of the camera poses are visually similar. In subsequent quantitative evaluations, we normalize the pose to reduce the impact of scale variations.

We use two types of metrics in the result evaluation: (1) Video quality: We use the Fréchet Inception Distance (FID)~\cite{FID} and Fréchet Video Distance (FVD)~\cite{fvd} to evaluate the quality of the generated videos. For FVD and FID metric reference, we utilize AnimateDiff with the same prompt to generate $800$ videos. Since all comparison methods are built on AnimateDiff, it is expected that the video quality after camera control should degrade minimally and align with that of AnimateDiff. (2) Pose error: In line with CameraCtrl~\cite{cameractrl}, we compute the Rotational Error $\operatorname{R_{error}}$ and Translational Error $\operatorname{T_{error}}$ between the estimated pose and the ground truth from RealEstate10K~\cite{realestate}. The formula for Rotational Error is detailed in Eq. \ref{rot}:
\begin{equation}
\operatorname{R_{error}}=\sum_{j=1}^n \arccos \frac{\left.\operatorname{tr}\left(\mathbf{R}_{est}^j \mathbf{R}_{g t}^{j \mathrm{~\top}}\right)\right)-1}{2},
\label{rot}
\end{equation}
where $\mathbf{R}_{est}^j$ represents estimated rotation matrix and $\mathbf{R}_{g t}^j$ represents ground truth rotation matrix for the $j$-th frame. The formula for calculating translational error is expressed as follows:
\begin{equation}
\operatorname{T_{error}}=\sum_{j=1}^n\left\|\mathbf{T}_{g t}^j-\mathbf{T}_{est}^j\right\|_2,
\label{tra}
\end{equation}
where $\mathbf{T}_{est}^j$ and $\mathbf{T}_{g t}^j$ respectively denote estimated translation matrix and ground truth translation matrix for $j$-th frame. Following ViewCrafter~\cite{viewcrafter}, we use DUSt3R~\cite{dust3r} to estimate the pose. For pose evaluation, we adjust poses to be relative to the first frame, followed by normalization on translation scale.

\subsection{Main Result}
\label{exp:mainresult}

We compare our results with MotionCtrl~\cite{motionctrl} and CameraCtrl~\cite{cameractrl}, which represent the state-of-the-art methods in camera-controlled video generation. CamTrol~\cite{CamTrol} achieves camera-controlled image-to-video generation, but it does not open source its model and code. Unlike MotionCtrl and CameraCtrl, which embed camera poses and fine-tune the diffusion model on large datasets, our method directly applies to the inference process of existing text-to-video diffusion models. It requires no training.

\noindent \textbf{Quantitative Evaluation.}
As shown in Table~\ref{tab:sota}, our method, without training, is comparable to and even surpasses other methods. In terms of video quality, our method achieves better FID and FVD scores than CameraCtrl. For pose estimation, our method achieves the lowest translation error ($\operatorname{T_{error}}$) and the second lowest rotational error ($\operatorname{R_{error}}$). We attribute the relatively higher rotational error to the reframe-by-inpaint process, which is more effective for modeling translational changes than for rotational changes.

For pose estimation, it is worth noting that each method has video instances where it fails to get pose.
 
We use the intersection $463$ out of $800$ samples of successful cases to compare the translational error and rotational error results.

\begin{table}[t]
  \caption{\textcolor{black}{\textbf{Quantitative Comparisons with training-based methods: MotionCtrl~\cite{motionctrl} and CameraCtrl~\cite{cameractrl}.} Our Method outperforms CameraCtrl on both video quality and camera pose accuracy. Meanwhile, our method achieves comparable or even better results against MotionCtrl on four evaluated metrics, without training. The best and second best results are marked with \textbf{bold} and \underline{underline}, respectively.}}
  \centering
  \setlength\tabcolsep{4pt}
  \begin{tabular}{c|ccc}
    \toprule 
    Method & MotionCtrl~\citep{motionctrl} & CameraCtrl~\citep{cameractrl} & Ours \\
    \midrule
    FID~$\downarrow$         & \underline{63.96}  & 92.40  & \textbf{60.18}\\
    FVD~$\downarrow$         & \textbf{468.29} & 531.69 & \underline{509.11}\\ 
    TransErr~$\downarrow$    & \underline{7.44}       & 8.39       & \textbf{5.52} \\
    RotErr~$\downarrow$      & \textbf{1.23}       & 2.57  & \underline{2.29} \\
    \bottomrule
  \end{tabular}
  \label{tab:sota}
\end{table}

\noindent \textbf{Visual Analysis.} 
As illustrated in \cref{fig:comparison}, while MotionCtrl maintains high video quality, it exhibits instances of semantic misalignment and slow motion. For example, in the left portion of row 2, no turtle appears in the ocean, and the waves remain nearly stationary across different frames. Conversely, CameraCtrl demonstrates lower video quality as shown in row 3. Our method achieves both high video quality and effective camera control in video generation without training.

\vspace{-1mm}
\subsection{Ablation Study}
\label{exp:ablation}

In the ablation study, we analyze the effect of time-aware point clouds, latent reframing steps, latent space harmonization, and different point cloud reconstruction models.

\noindent \textbf{Time-Aware \vs Static Point Cloud.} We compare two forms of point clouds: a time-static point cloud, which merges point clouds predicted from all images into a single global point cloud and renders the video from the target pose; and a time-aware point cloud, which assigns a temporal scale to the point cloud generated from each frame and considers only the corresponding temporal point cloud during target pose rendering.

\begin{figure}[t]
    \centering
    \includegraphics[width=1\linewidth]{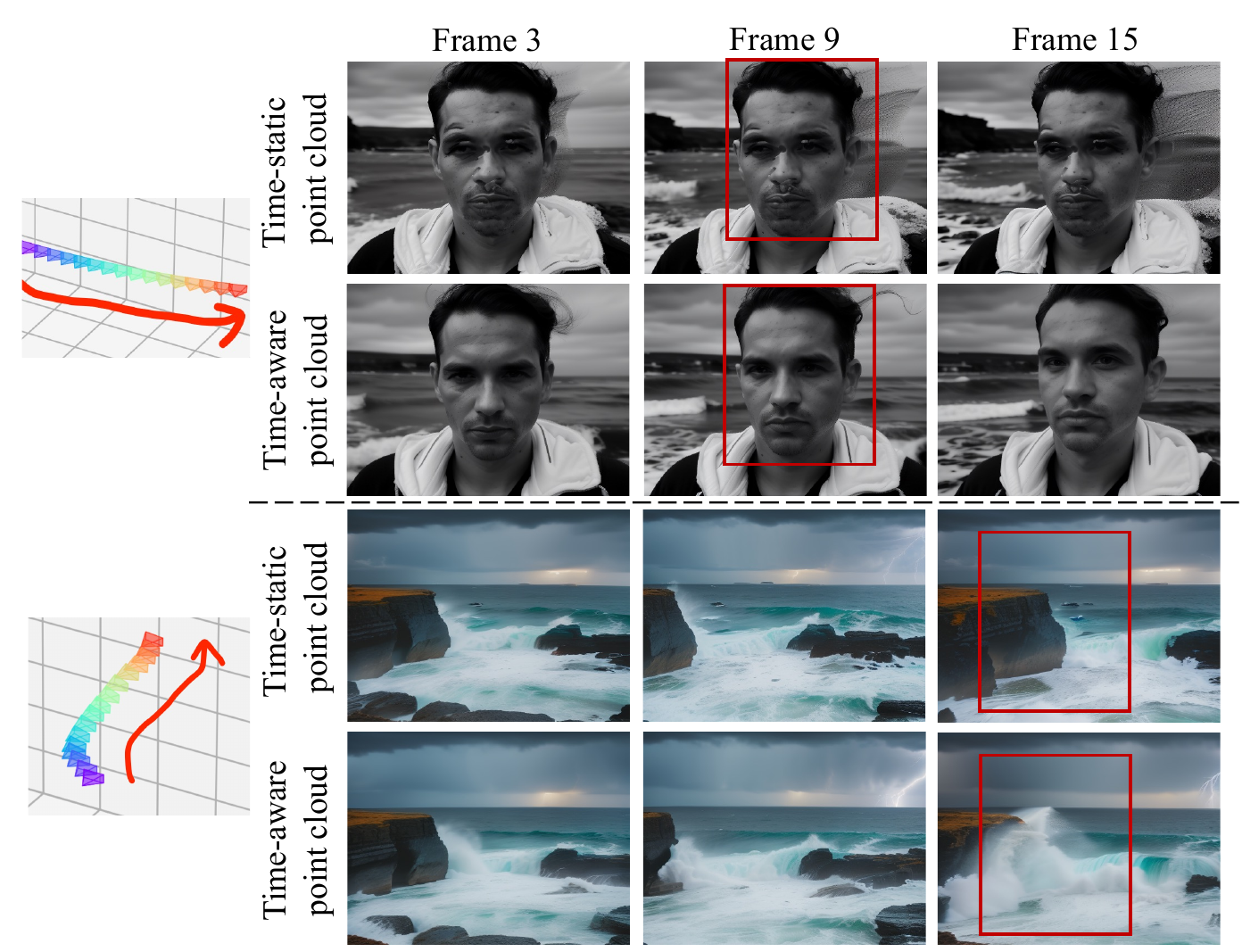}
    \caption{\textbf{Comparison between the time-aware and time-static point clouds.} Time-aware point cloud can capture more temporal dynamics of the video, For instance, the motion of the human face (row 1 and 2) and wave (row 3 and 4) are more prominent using time-aware point cloud, both are marked with \textcolor{red}{red} bounding boxes.}
    \label{fig:perframe_point}
\end{figure}

As shown in \cref{fig:perframe_point}, videos predicted using the single global point cloud appear static. Although minor movements can be observed through the video diffusion model (\eg, the small waves behind the human in the upper rows), the faces of the human remain static and exhibit noise, and the wave movement in the lower rows is minimal. In contrast, videos predicted using the time-aware point cloud retain motion information, such as changes in facial orientation in the upper rows and significant wave movement in the lower rows, while also achieving higher image quality, as highlighted by the red boxes in the figure. This demonstrates that the time-aware point cloud effectively captures and preserves the original dynamic information of the video diffusion model and accommodates corresponding camera control changes.

\begin{figure}[t]
    \centering
    \includegraphics[width=1\linewidth]{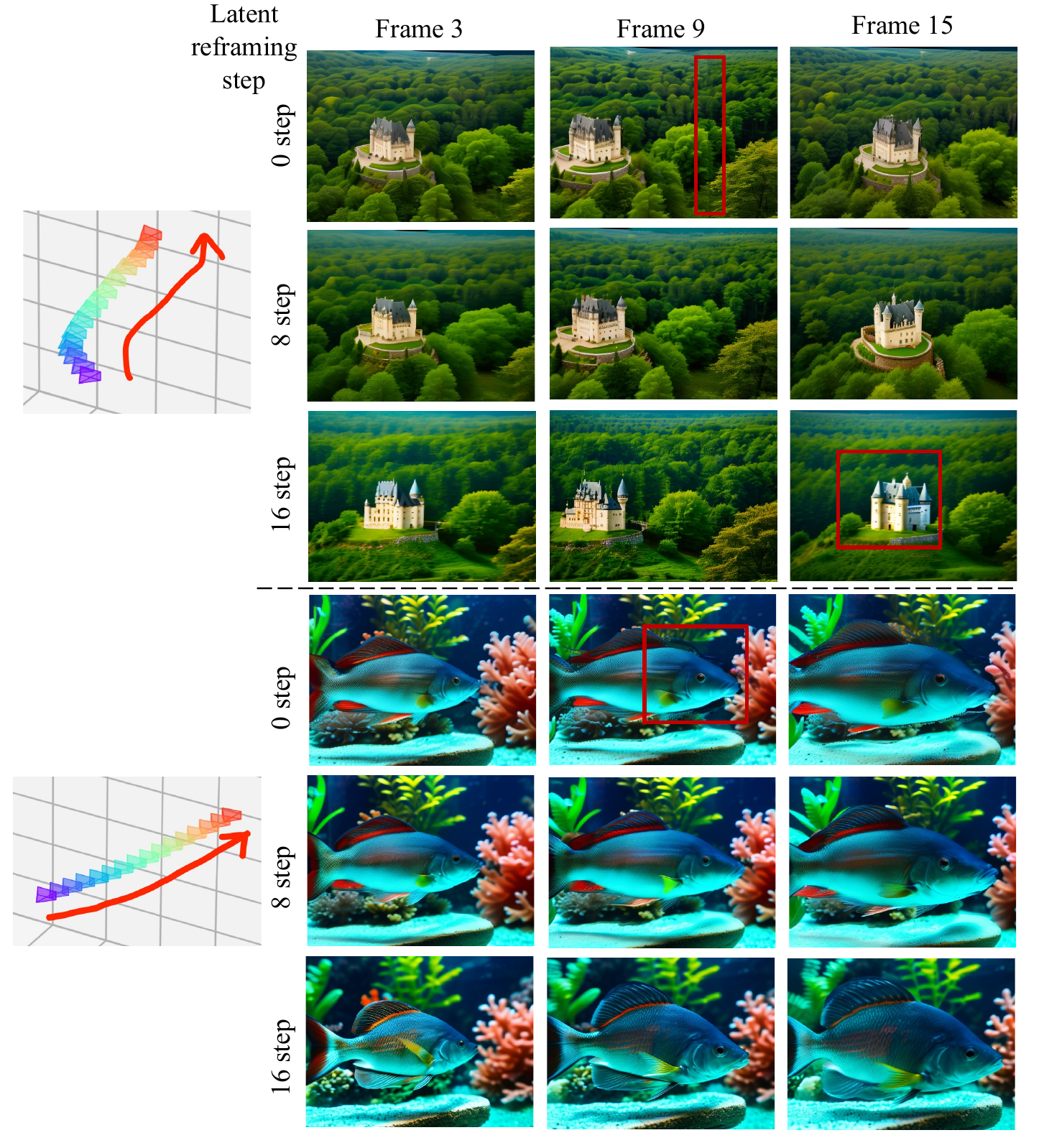}
    \caption{\textbf{Comparison between diffusion steps to apply \modelname}. Using diffusion step $8$ out of $25$ allows for the reconstruction of high-precision point clouds while leaving enough room for latent space inpainting and harmonization.}
    \label{fig:renderstep}
    \vspace{-3mm}
\end{figure}

\noindent \textbf{Diffusion Step Choice for Latent Reframing.}  The latent warping refers to the step in the denoising process of the diffusion model where the latent noisy video is warped to the target poses. We use a total of $25$ denoising steps in diffusion model inference, with step $25$ corresponding to the state of complete Gaussian noise and step $0$ indicating the fully denoised image. We consider three scenarios: performing latent warp at diffusion step $0$, $8$, and $16$.

When the latent warp step is set to 0, the video is fully denoised before pose warping. This approach benefits from the absence of noise, resulting in more accurate pose estimation for the points and images. However, this leads to potential disharmony during the latent inpainting process, as the denoising is applied exclusively to the masked region without integrating the entire image. This issue is evident in \cref{fig:renderstep}, where noticeable strip artifacts appear in the first and fourth rows. Conversely, when the latent warp step is set to 16, the image is at a high noise level, which compromises the accuracy of the reconstructed point cloud. Additionally, the generative capabilities of the video diffusion model overshadow the camera control information provided by the time-aware point cloud in denoising, leading to inconsistencies with the original video content. This is illustrated in the third rows of \cref{fig:renderstep}, where the final columns display content different from previous frames. Setting the latent warp step to 8 strikes a balance, maintaining a moderate noise level that allows for the reconstruction of high-precision point clouds while effectively filling in gaps to match the original video content. This results in harmonious and high-quality videos under new camera poses.

\noindent \textbf{Noise Reduction Strength for Latent Harmonization.} We investigated the impact of latent harmonization, achieved by reducing noise in known regions, on the generation quality of unknown regions. This was measured by decreasing the number of denoising steps, for instance, reducing from $15$ to $12$ steps represents a reduction of three denoising steps. 

\begin{figure}[t]
    \centering
    \includegraphics[width=1\linewidth]{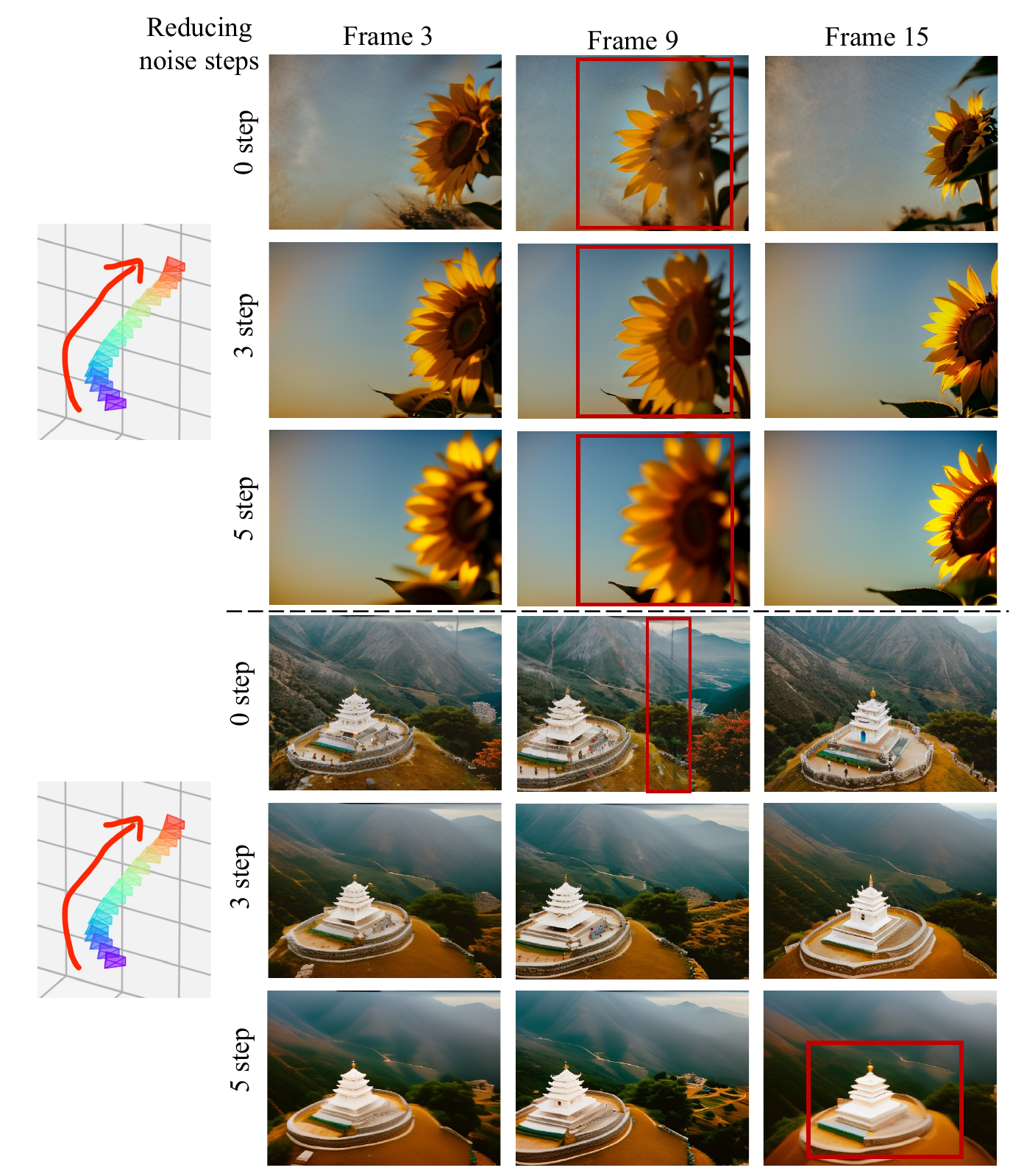}
    \caption{\textbf{Comparison between noise reduction steps.} Reducing the noise level of the known region by $3$ steps results in videos without significant blurriness or strip artifacts, reaching the best balance.}
    \label{fig:noisereduce}
    \vspace{-3mm}
\end{figure}

We tested reductions of $0$, $3$, and $5$ steps. As shown in \cref{fig:noisereduce}, when no noise reduction is applied to the known regions, the final video exhibits noticeable artifacts. In the first row, numerous noise points are observed around the sunflower, while in the fourth row, distinct banding artifacts appear in the middle of the image. This occurs because the denoising process in the known regions does not account for content generation in unknown regions. Conversely, when noise is reduced by $5$ steps, the large noise discrepancy between known and unknown regions leads to blurriness, as illustrated in the third and sixth rows. However, reducing noise by $3$ steps results in videos without significant blurriness or strip artifacts, achieving the best generation quality. These results indicate that appropriately reducing noise in known regions can significantly enhance the generation of unknown regions, leading to more harmonious videos.

\begin{figure}[t]
    \centering
    \includegraphics[width=1\linewidth]{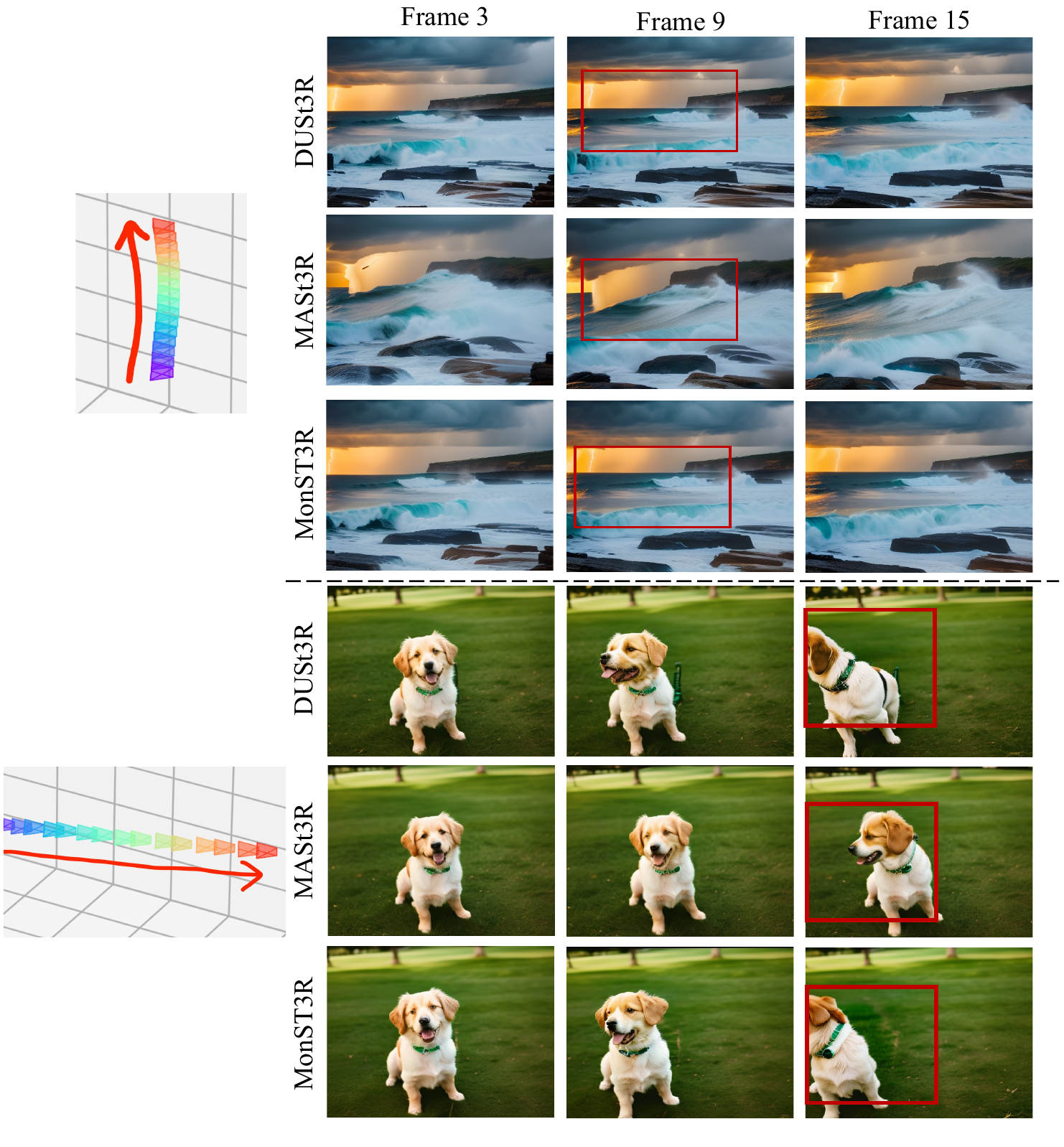}
    \caption{\textbf{Comparison between point cloud extraction methods.} MonST3R exhibits better performance in keeping fine details of the video.}
    \label{fig:differentmodel}
    \vspace{-3mm}
\end{figure}

\noindent \textbf{Choice of Point Cloud Reconstruction Model.} We evaluated three models for extracting time-aware point-cloud information from videos: DUSt3R~\cite{dust3r}, MASt3R~\cite{mast3r}, and MonST3R~\cite{monst3r}. As illustrated in \cref{fig:differentmodel}, both MonST3R and DUSt3R demonstrated comparable performance, successfully reconstructing high-quality point clouds suitable for rendering. Notably, MonST3R exhibited superior performance in keep fine details of videos. For instance, in the last column of the fourth row, DUSt3R produced an elongated neck for the dog, whereas MonST3R maintained a realistic appearance. This improvement can be attributed to MonST3R's training on dynamic datasets, which enhances its ability to perceive and process dynamic image sequences. Conversely, MASt3R's reconstructions were less accurate, particularly in pose estimation. This is evident from the significant pose errors in the second row and the failure to reach the target pose in the fifth row.

%% file: sec/5_conclusion.tex
\section{Conclusion}
This paper presents \modelname, a method for enabling camera control in video diffusion models without requiring additional training. \modelname operates during the denoising process with two key steps: a latent reframing step that adjusts the video’s latent code using time-aware point clouds and camera poses, and a latent space rehabilitation step that inpaints occluded regions and harmonizes the latent code for high-quality video generation. Experimental results show that \modelname achieves comparable or superior camera control precision and video quality compared to training-based methods. This work highlights the effectiveness of leveraging 3D information to enhance 2D video generation.

%% file: sec/6_supp.tex
\clearpage
\appendix



\section{More Results}
\label{moreresult}
In this section, we present additional results of \modelname in camera control for video generation to validate its applicability. These results include both basic camera poses (translational and rotational) and various video styles.

\begin{figure*}[h]
    \centering
    \includegraphics[width=0.95\linewidth]{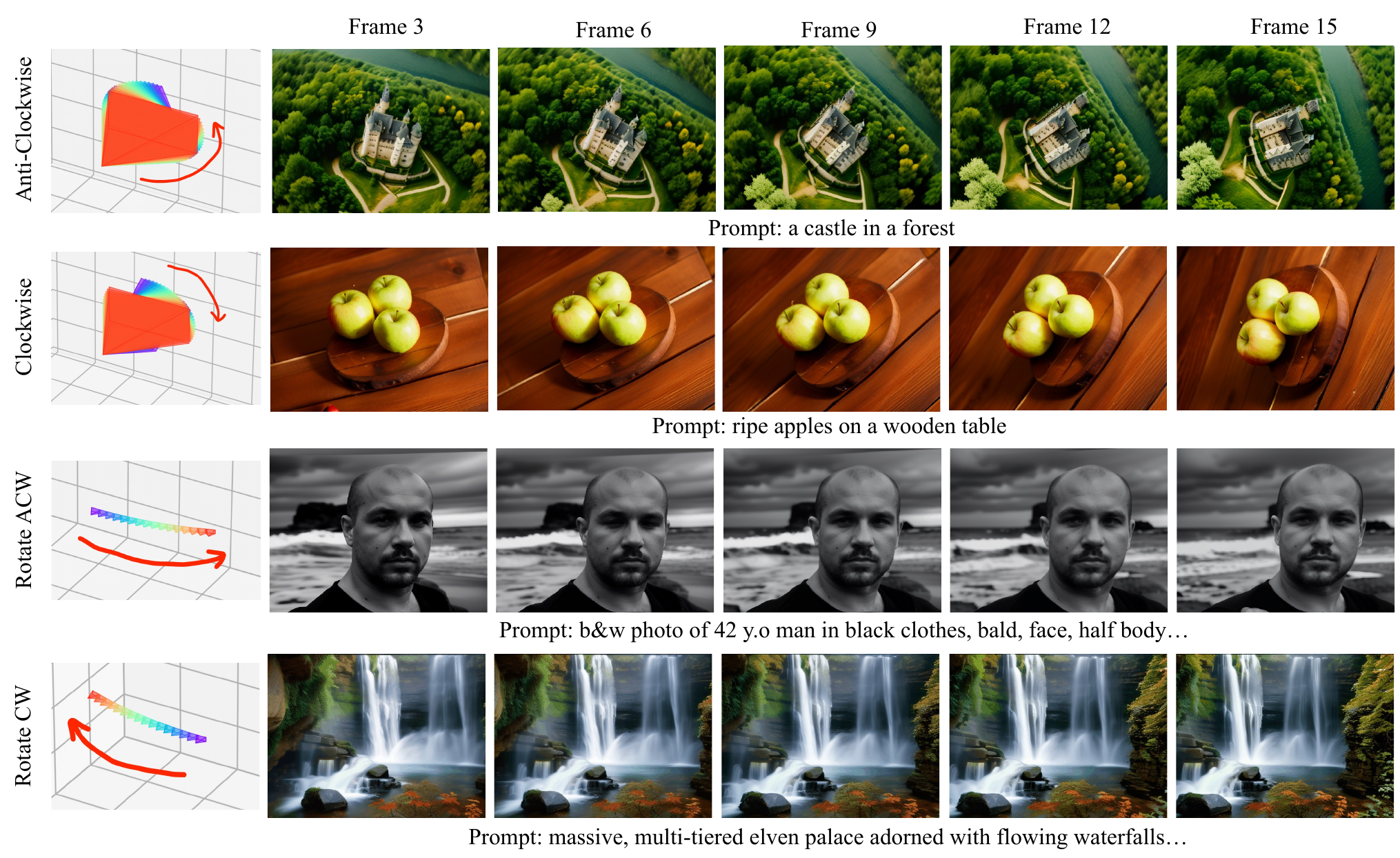}
    \caption{\textbf{Results of \modelname camera control for video generation based on rotational basic poses.}}
    \label{fig:basicpose2}
    \vspace{-3mm}
\end{figure*}

\begin{figure*}[t]
    \centering
    \includegraphics[width=1\linewidth]{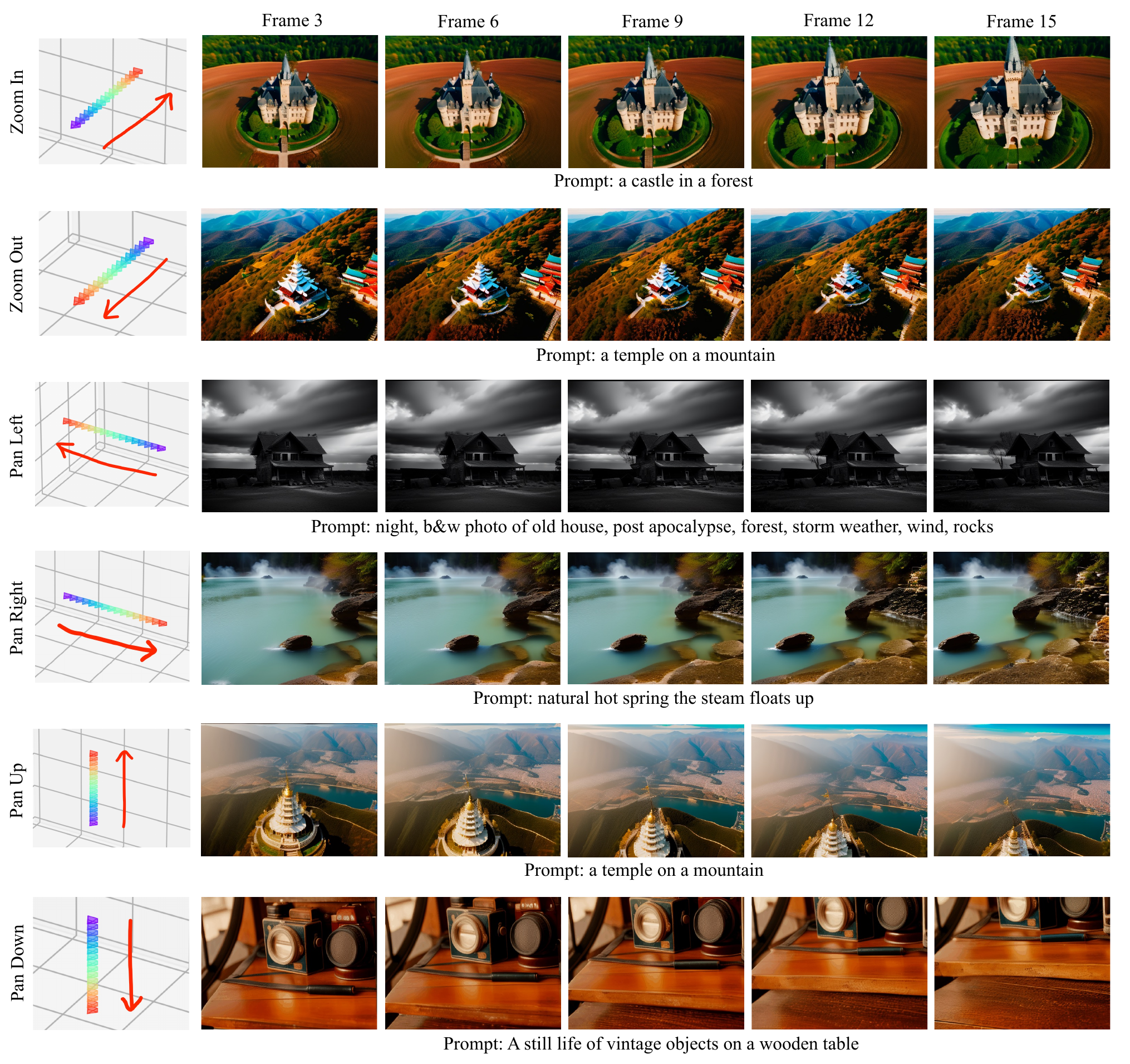}
    \caption{\textbf{Results of \modelname camera control for video generation based on translational basic poses.}}
    \label{fig:basicpose1}
\end{figure*}

Following MotionCtrl~\cite{motionctrl}, the translational basic camera poses comprise six types: zoom in, zoom out, pan left, pan right, pan up, and pan down, as illustrated in ~\cref{fig:basicpose1}. The rotational basic camera poses include four types: the camera's own rotations (clockwise and counterclockwise) and the camera's rotations around an object (rotate clockwise and counterclockwise), as shown in ~\cref{fig:basicpose2}. These results indicate that \modelname can execute various basic camera controls without requiring any training.

\begin{figure*}[t]
    \centering
    \includegraphics[width=1\linewidth]{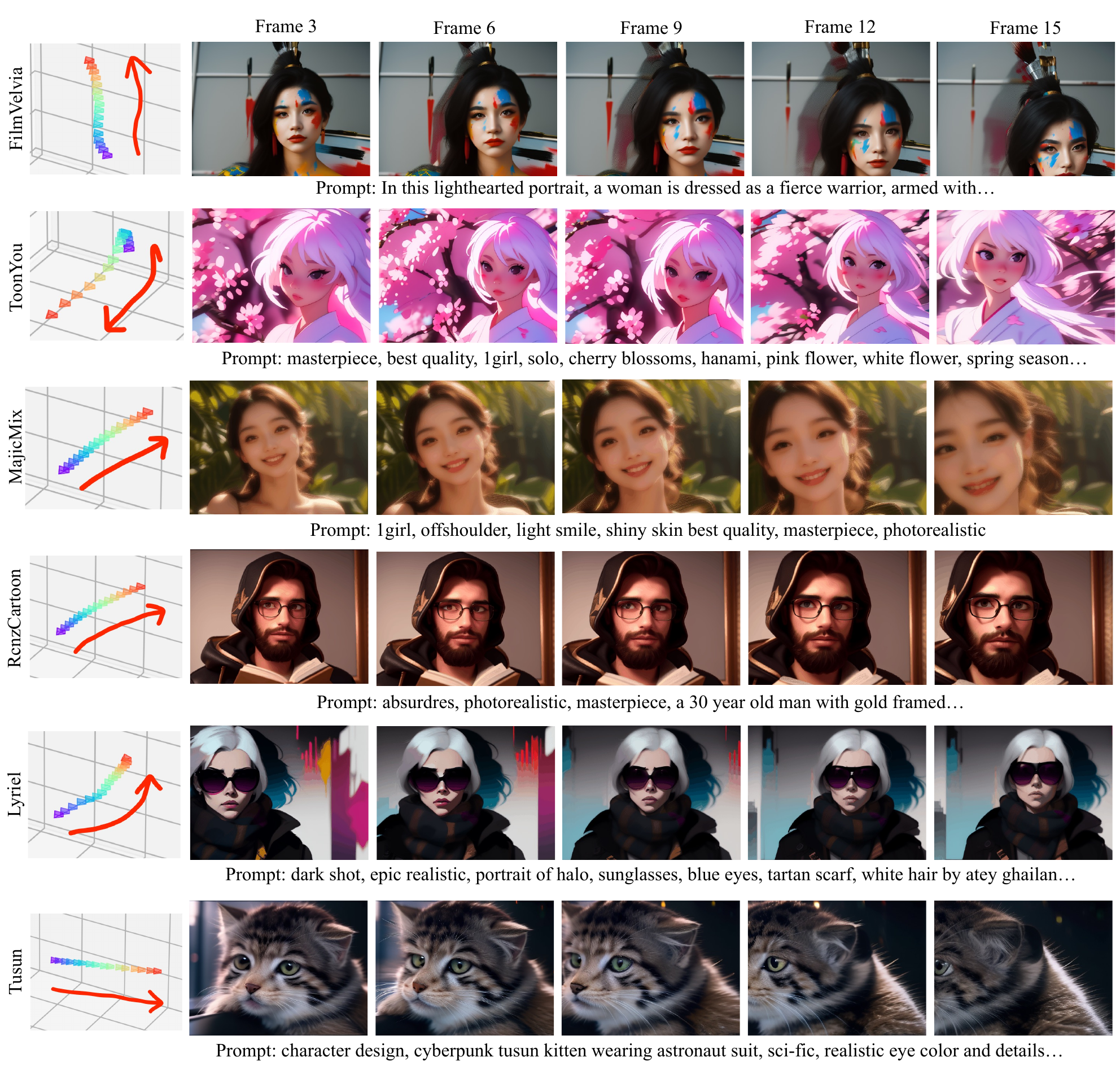}
    \caption{\textbf{Results of \modelname complex camera control for various video style.}}
    \label{fig:style}
\end{figure*}

For video generation in different styles, we follow AnimateDiff~\cite{animatediff} and apply complex camera pose control across six different video generation styles: FilmVelvia, ToonYou, MajicMix, RcnzCartoon, Lyriel, and Tusun, as shown in \cref{fig:style}. These results demonstrate that \modelname can effectively manage camera control for video generation across a wide range of styles, highlighting its versatility and suitability.

\begin{figure}[h]
    \centering
    \includegraphics[width=1\linewidth]{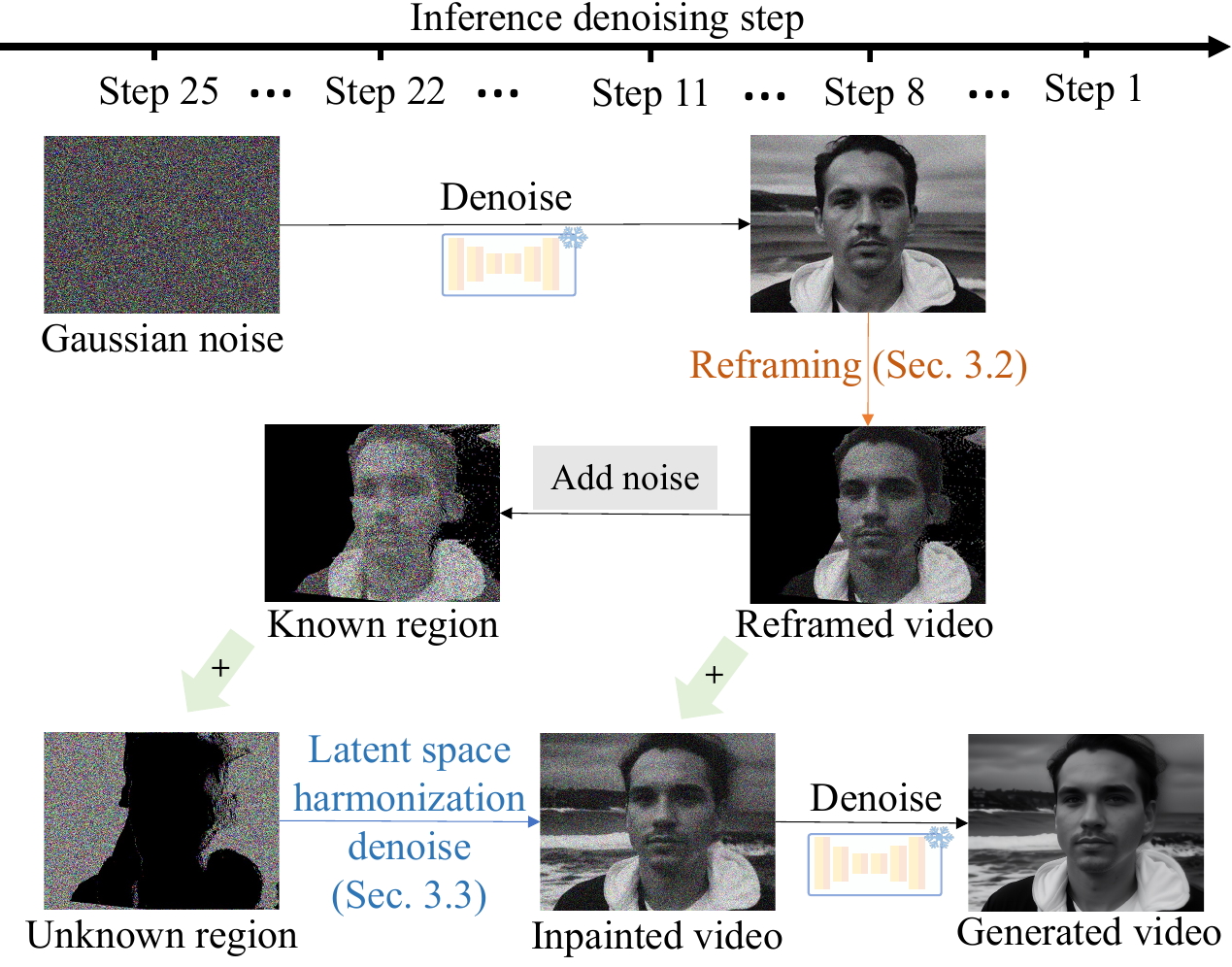}
    \caption{\textbf{Denoising time line of \modelname.} Here, the diffusion step for latent reframing is set to 8, and the noise reduction step is set to 3.}
    \label{fig:timeline}
    \vspace{-3mm}
\end{figure}

\section{Denoising Process Details}
\vspace{-3mm}
\label{methoddetail}
To provide a clearer explanation of the \modelname inference denoising process, we present the denoising timeline in \cref{fig:timeline}.

\begin{enumerate}
    \item The process begins with fully Gaussian noise at the $25^{\rm th}$ step. Denoising is conducted using the video diffusion model until reaching the predetermined latent reframing step, which is step 8.
    \item Latent reframing, as described in Sec.~3.2, is then applied to generate the reframed video at the target camera pose. At this point, holes caused by occlusions are present, and the regions are differentiated into known and unknown regions.
    \item The noise addition process starts. As outlined in Sec.~3.3, the noise level in the known region is set to be $3$ steps lower than that in the unknown region. For instance, when the unknown region is at the $25^{\rm th}$ denoising step, the known region is at the $22^{\rm th}$ denoising step.
    \item Denoising of the unknown region proceeds. At this stage, the input to the denoising network combines unknown and known regions, as described in Eq.~4 of the main paper. The output of the denoising network updates the unknown region for the next step, while the known region for the next step is obtained by adding corresponding noise to the reframed video.
    \item When the unknown region reaches the $11^{\rm th}$ denoising step, the known region has already reached the $8^{\rm th}$ step, which is the step for latent reframing. After this, the known region is no longer merged. The denoising process continues normally, updating the entire video until the final step.
\end{enumerate}

\section{Hyperparameter Settings}
\vspace{1mm}
\label{paradetail}
\cref{tab:hyperparameters} further details the parameters employed in \modelname.

\begin{table}[h]
\centering
\caption{\bf Detail hyperparameter settings.}
\setlength\tabcolsep{8pt}
\small
\begin{tabular}{cc}
\toprule
\textbf{Parameter} & \textbf{Value} \\
\midrule
Video frames & $16$ \\
Spatial resolution & $512{\times}384$ \\
DDIM denoising steps & $25$ \\
Classifier-free guidance scale & $7.5$ \\
Diffusion step for \modelname & $8$ \\
Noise reduction step & $3$ \\
Point map prediction model & MonST3R~\cite{monst3r} \\
Sliding window size & $3$ \\
Globally aligned points optimization steps & $300$ \\
\bottomrule
\end{tabular}
\label{tab:hyperparameters}
\vspace{-3mm}
\end{table}

\section{Limitation}
\vspace{1mm}
\label{limitaion}
While our method effectively enables video generation with camera control without requiring training, it does have certain limitations. For instance, when generating with specific camera poses, especially those involving large movements, can result in visual inconsistencies and artifacts due to extensive unknown regions. Additionally, inaccuracies in the point cloud estimation model and pose estimation can significantly compromise the quality of camera control, leading to errors. Future research will focus on analyzing and mitigating the failure rates associated with these challenging camera poses. 

